Corbion

intrepid

# Accelerating the Development of PLGA *In Situ* Forming Depots Through AI-Driven Multi-Objective Optimization

*A Corbion-Intrepid Labs case study in balancing drug loading, sustained release, and injectability*

Pauric Bannigan,[†] Siddarth Chandrasekaran,[†] Brigitte A. G. Lamers,[‡] Inge Hermsen,[‡] Gary Tom,[†] Riley J. Hickman,[†] Bahar Yeniad,[‡] Morgan Fox,[‡] Christine Allen[†]

*†Intrepid Labs Inc., 140 - 101 College St, Toronto, ON M5G 1L7. [E]: christine@intrepidlabs.ai*
*‡Corbion N.V., Arkelsedijk 46, Gorinchem, 4206 AC, [E]: morgan.fox@corbion.com*

## ABSTRACT

Developing long-acting injectable formulations requires the simultaneous optimization of drug loading, release kinetics, viscosity, injectability, stability and other objectives. To navigate this multidimensional space, Corbion and Intrepid combined Corbion's diverse PURASORB® bioresorbable polymer library with Intrepid Labs' proprietary AI algorithm (ANDROMEDA 1) to develop *in situ* forming depots for a therapeutic peptide. Over approximately 15 weeks, 181 unique formulations spanning drug loadings of 6–12% w/w were prepared and characterized through broad design-space mapping and targeted multi-objective optimization. Four lead candidate formulations were identified at 6%, 9%, and 12% w/w drug loading. Each met the predefined viscosity and injectability criteria while providing distinct 30-day *in vitro* release profiles. The study evaluated polymers spanning a broad range of molecular weights, including commercially available PURASORB® grades and new polymers under development by Corbion to expand its polymer toolbox. ANDROMEDA 1 identified that polymers with intermediate molecular weights provided a favorable balance between sustained release and solution viscosity. Together, these findings demonstrate how integrated polymer expertise and AI-driven optimization can rapidly identify differentiated formulation candidates, focus the development space, and establish a strong data-driven foundation for further optimization and *in vivo* evaluation.

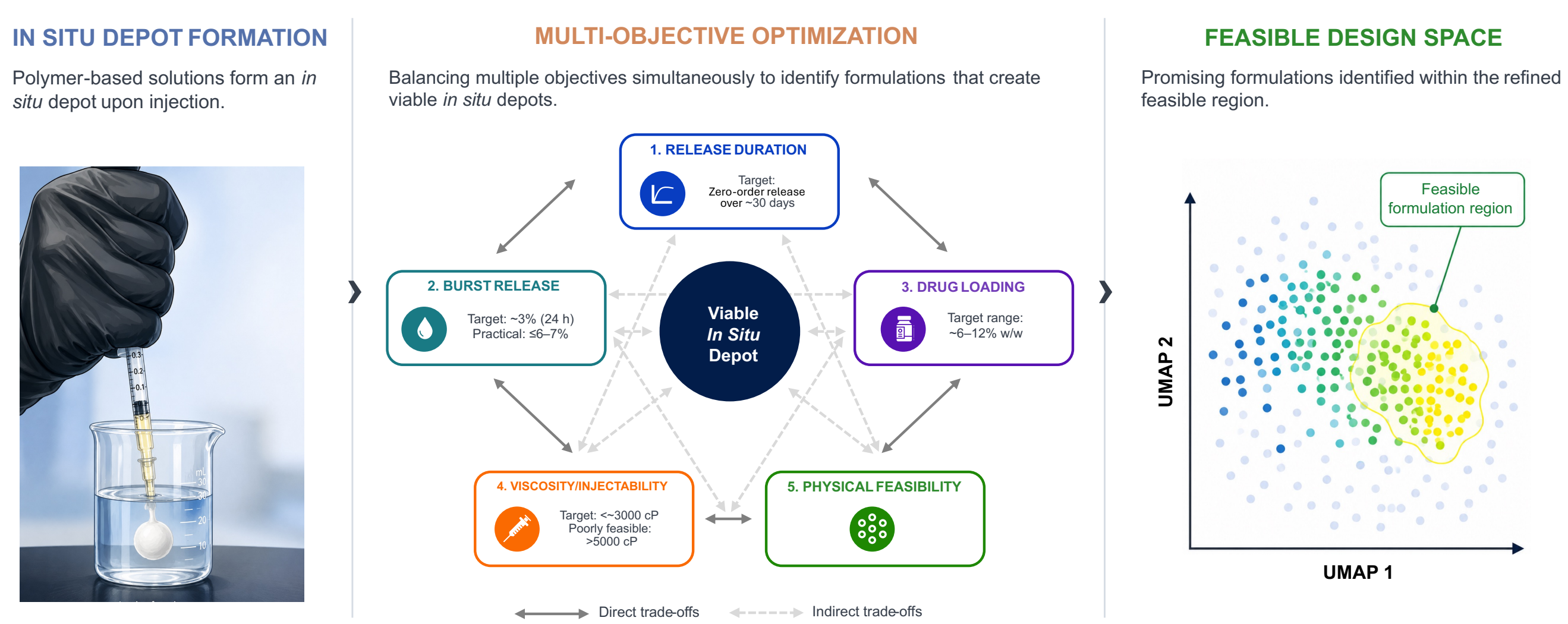

**Figure 1. Iterative, model-guided optimization of *in situ* forming depots.** The workflow balances multiple formulation objectives including release duration, burst release, drug loading, viscosity and injectability, and physical feasibility. Experimental results from each batch are used to refine the model and guide subsequent experiments toward the most promising regions of the design space. By continuously learning from prior experiments, this iterative approach reduces the number of formulations that must be tested, improves experimental efficiency, and progressively identifies a refined feasible region in which performance objectives are jointly satisfied.

## INTRODUCTION

Patient-centered care considers not only clinical efficacy, but also how well a therapy fits the needs, preferences, and circumstances of the people who use it. Across serious mental health conditions, substance use disorders, HIV, cancer and endocrine disorders, long-acting injectables (LAIs) can expand treatment choice, reduce the burden of frequent dosing, and support continuity of treatment.

Historically, bioresorbable polymer-based LAIs have relied on preformed PLGA microparticles or microspheres. Although clinically validated, their multistep manufacture can slow development and complicate scale-up.[1] *In situ* forming depots offer a promising alternative by forming a semi-solid drug-containing polymer depot after injection of the polymer solution. Products including ELIGARD® for advanced prostate cancer, SUBLOCADE® for opioid use disorder and PERSERIS®, UZEDY®, and RISVAN® for schizophrenia demonstrate the clinical utility of *in situ* forming depots across multiple therapeutic areas.

However, avoiding microsphere manufacture does not eliminate practical constraints at the point of use. ELIGARD®, PERSERIS®, and RISVAN® are supplied as dual-syringe systems that require mixing immediately before administration. ELIGARD® and PERSERIS® each require 60 back-and-forth mixing cycles, while RISVAN® requires at least 100 rapid alternating pushes and must be administered within 15 minutes of reconstitution. SUBLOCADE® and UZEDY® are supplied as ready-to-use, single-syringe presentations, but still have handling and administration requirements related to temperature and viscosity. These examples demonstrate why formulation optimization must balance drug loading and release performance with preparation simplicity, viscosity, injectability, and reliable dose delivery.

Realizing the full potential of *in situ* forming depots requires navigating a multidimensional design space spanning polymer chemistry, molecular weight, architecture, end group chemistry, and concentration; solvent identity and level; drug loading; and the identity and concentration of any additives. The number of possible combinations of these variables quickly becomes too large to evaluate experimentally, and interactions among them must be balanced against competing objectives. These include release duration, initial burst release, viscosity, injectability, stability, manufacturability and cost of goods (**Figure 1**).

Machine learning (ML) integrated with automated experimental workflows provides a practical way to navigate large design spaces efficiently. ML-guided multi-objective optimization directs experimentation toward formulations that balance a more complete target product profile, rather than optimizing individual attributes in isolation. As experimental data accumulate, models learn composition-performance relationships, quantify predictive uncertainty and help identify the boundaries within which acceptable performance is maintained, providing an early indication of formulation robustness.

This white paper describes how Corbion's commercially available PURASORB® bioresorbable polymers and innovative PURASORB® polymer materials under development were combined with Intrepid Labs' VALIANT™ automated experimental platform and ANDROMEDA 1 ML algorithm to develop and optimize a one-month *in situ* forming depot containing a therapeutic peptide.

The case study illustrates how combining advanced polymer science with AI-guided experimentation can accelerate LAI development while generating the formulation understanding needed to design robust, scalable products.

### PURASORB Polymer Toolbox

The PURASORB® bioresorbable polymer portfolio provides a versatile set of materials for tailoring the performance of *in situ* forming depot formulations. However, commercially available polymers do not always provide the full range of attributes needed to achieve a desired product profile. This contributes to persistent formulation challenges and may be one factor underlying the relatively limited number of FDA-approved *in situ* forming depot products.[2] Two key challenges associated with poly(D,L-lactide-co-glycolide) (PLGA)-based systems are maintaining viscosity and injection force within acceptable limits while controlling initial burst release.[3]

Expanding the available range of polymer attributes could help address these challenges. Corbion's advanced synthesis capabilities have enabled the development of innovative polymers that broaden the current formulation toolbox for *in situ* forming depots. In this study, these polymers under development were evaluated alongside commercially available PURASORB® polymers. The resulting 15-polymer library included PLGA copolymers with lactide-to-glycolide molar ratios of 50:50 and 75:25 and poly(D,L-lactide) (PDL) homopolymers, spanning a range of molecular structures and inherent viscosities. Together, these materials enabled systematic evaluation of how polymer attributes, including composition and properties, influence depot formation, peptide release kinetics, and physical formulation properties, such as viscosity.

These polymer attributes influence formulation performance through different mechanisms. Increasing polymer molecular weight can support more sustained release but typically increases formulation viscosity and may compromise injectability. The lactide-to-glycolide ratio of the copolymer provides an additional design lever by influencing copolymer hydration and degradation rate. End group chemistry represents another potential design variable, although it did not emerge as a meaningful differentiator among the materials evaluated in this campaign and will be investigated further in future studies. Collectively, the library provides a broad material design space for balancing release performance and practical formulation requirements.

### VALIANT™ for Multi-Objective Optimization

Intrepid Labs' VALIANT™ AI-driven experimental platform was used to systematically prepare and characterize *in situ* forming depot formulations across the Corbion polymer design space. Formulations were evaluated against a predefined target product profile encompassing release duration and trajectory, 24-hour burst release, drug loading, viscosity, injectability, and physical feasibility (**Table 1**). Defining these criteria at the outset allowed formulation performance to be assessed against the complete target product profile rather than release behavior alone.

**Table 1.** Target product profile used for multi-objective optimization (MOO).

| Attribute | Development objective |
| --- | --- |
| Release duration and profile | Sustained, near-zero-order release over approximately 30 days without secondary burst events |
| Initial burst | Ideal target of approximately 3% at 24 hours; practical upper threshold of approximately 6–7% |
| Drug loading | 6–12% w/w |
| Formulation viscosity | Preferred below approximately 3,000 cP; formulations above 5,000 cP considered poorly feasible |
| Physical performance | Homogeneous formulation, reproducible depot formation, and absence of phase separation |
| Injectability | Compatible with preparation and passage through a 25 G needle |

Experimental results were incorporated into ANDROMEDA 1, Intrepid's model-guided optimization engine, after each round of testing. ANDROMEDA 1 learns relationships between formulation composition and measured performance, estimates performance and uncertainty across the remaining design space, and prioritizes subsequent formulations for experimental evaluation. New data are then returned to the model, progressively refining the formulation landscape and concentrating experimentation within regions most likely to satisfy the target product profile.

This differs from conventional sequential development, in which individual attributes may be optimized before downstream constraints are considered. For *in situ* forming depots, these objectives are strongly coupled: design variable settings that improve release control may increase formulation viscosity or compromise injectability, while changes in polymer concentration, molecular weight, or drug loading can alter early burst and longer-term release, as well as impact formulation viscosity and injectability. The resulting trade-off is intrinsic to the system: formulations with similar 24-hour release can occupy substantially different viscosity regimes. Multi-objective optimization therefore seeks a feasible region in which release and practical administration requirements are satisfied concurrently, rather than a single optimum defined by one endpoint.

## RESULTS

The experimental campaign revealed a clear trade-off between release control and formulation practicality. Across the polymer library, increasing polymer molecular weight and polymer loading generally increased formulation viscosity, in some cases substantially, while also influencing initial burst and longer-term release. Higher viscosity may slow drug diffusion from the polymer depot and support sustained release, but it also increases resistance during formulation handling and injection. Conversely, lower-viscosity formulations are generally easier to prepare and administer but do not necessarily provide sufficient control over early or sustained release. This multidimensional relationship is evident in **Figure 2**, where formulations with similar 24-hour release values span a broad viscosity range. The selected lead formulations occupy the preferred region of this landscape, demonstrating that controlled early release can be achieved alongside practical viscosity across the tested drug loading range of 6–12% w/w.

These trends reflect the coupled effects of polymer properties and formulation composition on depot formation and drug release. Polymer molecular weight and concentration influence chain entanglement and solution viscosity, while polymer composition and solvent exchange affect phase inversion for depot formation, drug entrapment, water ingress, and subsequent release. Consequently, formulations that effectively suppress initial burst may release the drug too slowly at later time points, while changes intended to accelerate longer-term release may compromise burst control or injectability. The formulation objective is therefore not to maximize any single material property, but to identify a polymer and composition range that balances early drug retention, sustained release, and practical administration.

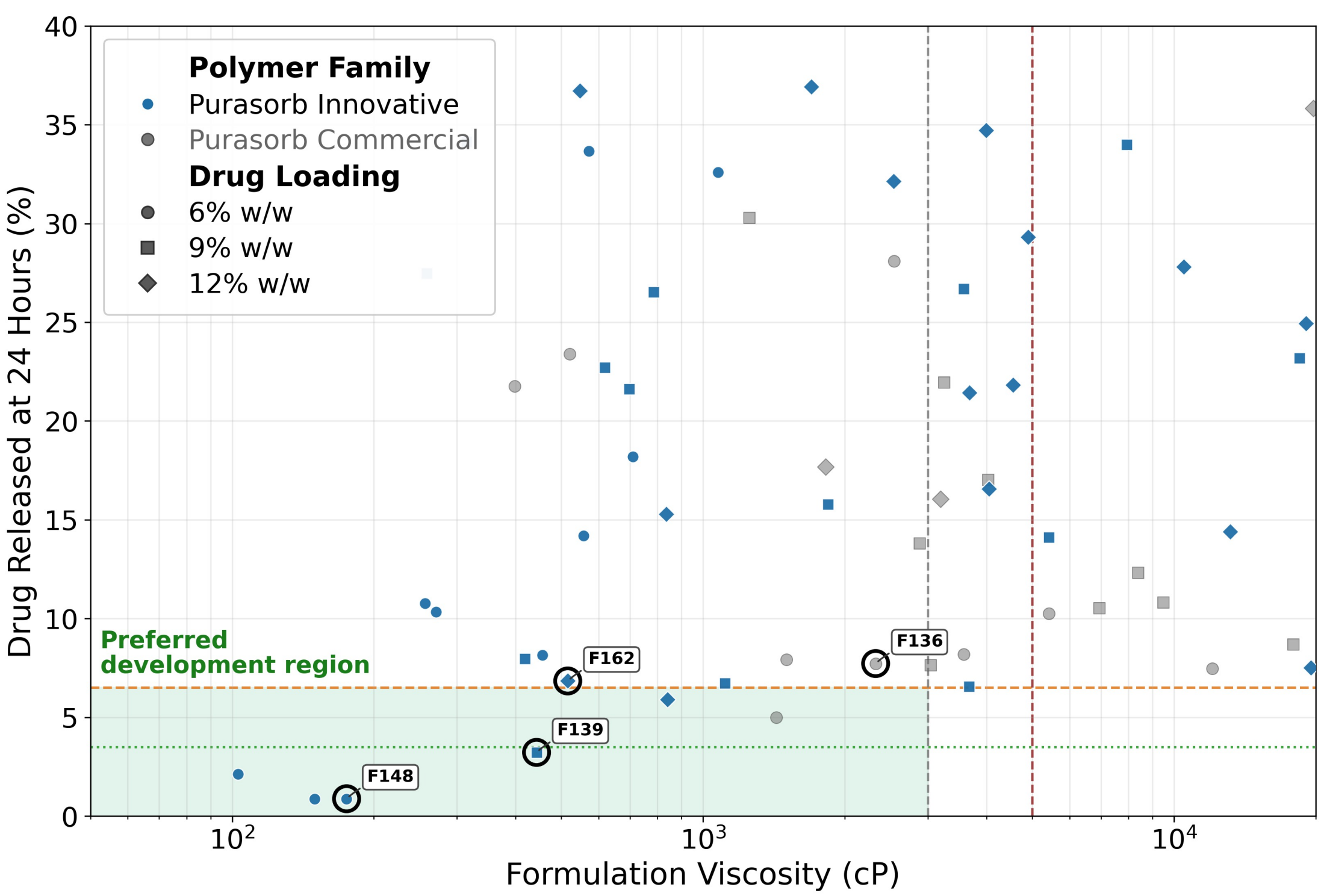

**Figure 2. Relationship between 24-hour burst release and viscosity during multi-objective optimization of *in situ* forming depots containing peptide.** The shaded region represents the preferred design space, defined by viscosity ≤3,000 cP and 24-hour burst release of approximately 6–7%. The vertical dashed line at 5,000 cP marks the threshold above which formulations are considered to have limited practical feasibility. Circled lead formulations (F136, F139, F148, and F162) span drug loadings of 6%, 9%, and 12% w/w. Marker color denotes polymer family, and marker shape denotes drug loading.

As shown in **Figure 2**, the preferred development region was populated predominantly by Corbion's innovative polymers. Most formulations meeting both the viscosity and burst release criteria, and three of the four lead candidates, incorporated innovative rather than commercially available materials. Access to these chemistries extended the accessible performance range beyond what the commercial portfolio alone supported, allowing sustained release to be achieved at viscosities suitable for administration through a 25 G needle. Within the library evaluated in this study, polymers with intermediate molecular weights provided the most favorable balance between sustained release and practical viscosity.

To identify the variables most strongly associated with formulation performance, the complete dataset was evaluated using model-interpretability analysis. **Figure 3** highlights the dominant formulation drivers for the two principal development objectives, 24-hour burst release and formulation viscosity, while the expanded partial-dependence analysis across all modeled objectives is provided in **Figure S1**. Polymer loading was the dominant driver of 24-hour burst release, with increasing polymer content associated with a marked reduction in burst. In contrast, higher residual monomer content and higher PLGA molecular weight were associated with increased burst release. Formulation viscosity was governed primarily by solvent fraction, decreasing sharply as solvent content increased, while higher molecular weight and higher intrinsic viscosity of the polymer were associated with progressively higher viscosity.

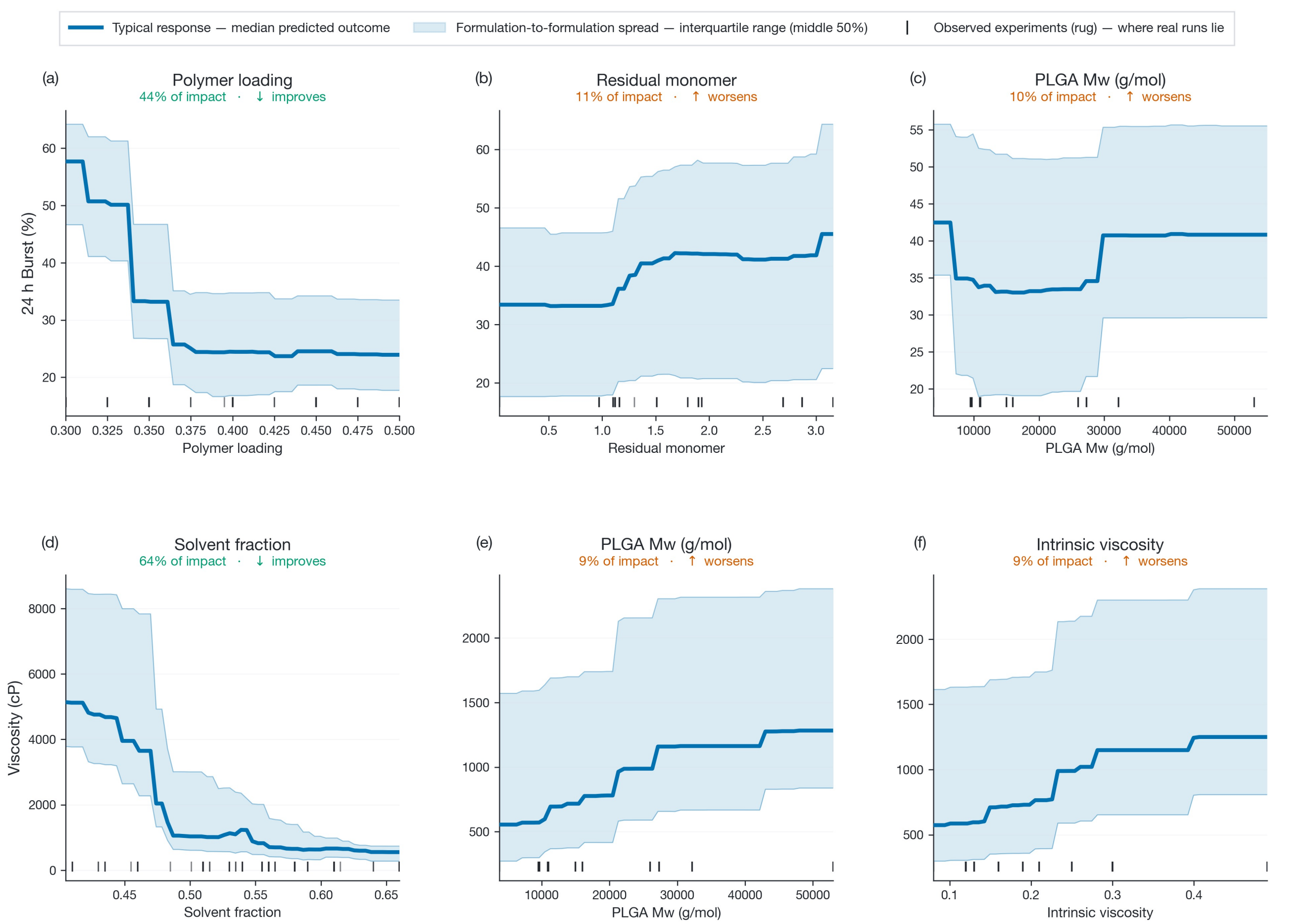

**Figure 3. Partial-dependence analysis of the key formulation drivers of 24-hour burst release and viscosity.** For each objective, the three most influential formulation attributes were varied across their 5th–95th percentile range while all other attributes retained their experimentally observed values. The solid line shows the median model-predicted response, and the shaded region shows the interquartile range across formulations. Rug marks denote the experimental coverage of each attribute, indicating where modeled trends are most strongly supported by measured data. Lower values are preferred for both objectives; downward trends therefore indicate improved performance, while upward trends indicate worsening performance. Percentages above each panel represent the attribute's relative contribution to the total mean absolute SHAP importance for that objective.

The expanded analysis in **Figure S1** showed that several of these formulation attributes also influenced deviation from zero-order release and overall multi-objective performance, with polymer loading and solvent fraction emerging as particularly important across multiple objectives. Several relationships were nonlinear, indicating threshold regions in which relatively small changes in composition or polymer properties produced substantial changes in formulation performance. Collectively, these results highlight the trade-offs governing depot performance and identify practical formulation ranges for balancing drug release and viscosity. They also provide insight into how polymer composition and molecular properties influence formulation behavior, supporting the rational design and accelerated development of new polymer materials for *in situ* forming depots.

Beyond identifying the formulation attributes associated with performance, the completed experimental dataset was also used to retrospectively evaluate the efficiency of alternative experiment-planning strategies. An oracle model, trained on the accumulated formulation and performance data and building on previously demonstrated machine-learning approaches for predicting the performance of polymeric long-acting injectables,[4] was used as a virtual laboratory to predict the outcomes of formulations proposed by ANDROMEDA 1, random search, and a conventional design of experiments (DoE) strategy. Repeated simulated campaigns therefore allowed each approach to explore the same underlying formulation landscape without physically repeating the experimental work, with performance assessed by the ability to identify a formulation meeting the predefined lead criteria for both 24-hour burst release and viscosity.

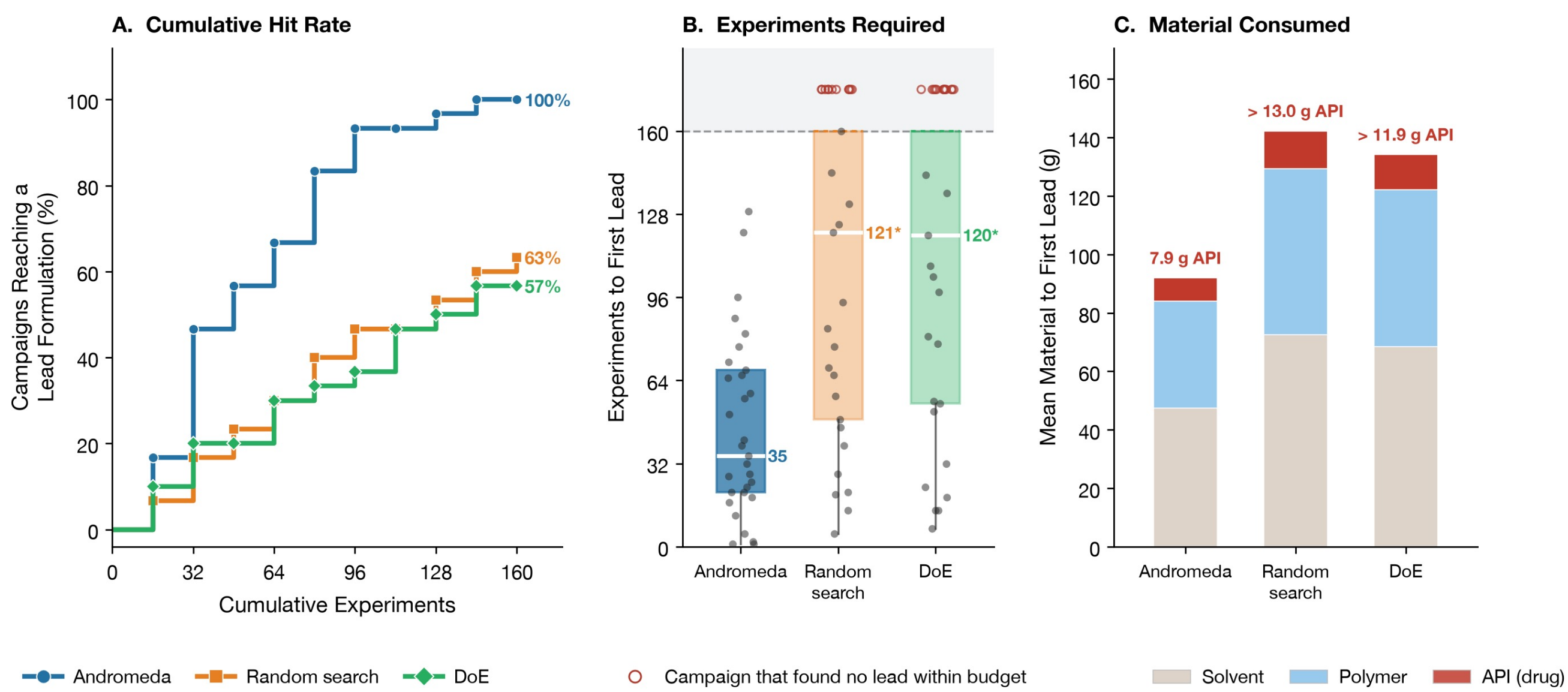

**Figure 4. Retrospective benchmarking of lead formulation discovery and material requirements.** An oracle model derived from the completed experimental dataset was used as a virtual laboratory to simulate formulation outcomes for experiments selected by ANDROMEDA 1, random search, and a conventional design of experiments (DoE) strategy. Each strategy was run as 30 independent campaigns of 160 simulated experiments conducted in batches of 16. A campaign was scored as successful on sampling any of the four formulations identified as leads in the completed experimental campaign (F136, F139, F148, F162). **(A)** Percentage of campaigns having reached a lead formulation as a function of cumulative simulated experiments. **(B)** Number of experiments required to reach the first lead formulation. Campaigns finding no lead within the 160-experiment budget are shown as open circles above the budget line, so medians are taken over all 30 campaigns; an asterisk denotes an upper quartile lying beyond the budget, drawn as an open box. **(C)** Estimated cumulative solvent, polymer, and peptide requirements to reach the first lead formulation, averaged over the campaigns that reached one. The analysis represents a retrospective computational comparison and not three independently executed experimental campaigns.

Clear differences were observed in both the probability and efficiency of identifying a lead formulation (**Figures 4A and 4B**). ANDROMEDA 1 progressively concentrated experiments within higher-performing regions of the design space and identified at least one lead formulation in 100% of simulated campaigns within the experimental budget. In comparison, only 63% of random-search campaigns and 57% of DoE campaigns identified a lead over the same search window. The difference was also reflected in the number of experiments required, after approximately 35 experiments with ANDROMEDA 1, compared with approximately 121 and 120 experiments for random search and DoE, respectively. Several random-search and DoE campaigns failed to identify a lead within the available experimental budget, further highlighting the greater consistency of the model-guided approach.

This improvement in experimental efficiency observed with ANDROMEDA 1 translated directly into lower material requirements (**Figure 4C**). Mean material consumption to identification of the first lead was approximately 92 g for ANDROMEDA 1, compared with approximately 141 g for random search and 134 g for DoE. Importantly for peptide formulation development, this included substantially lower peptide consumption: approximately 7.9 g for ANDROMEDA 1, compared with >13.0 g for random search and approximately 11.9 g for DoE, together with corresponding reductions in polymer and solvent requirements. Although this analysis represents a retrospective computational comparison rather than a prospective head-to-head experimental study, it illustrates the practical value of using accumulated formulation knowledge to guide subsequent experimentation. By preferentially selecting experiments with greater expected value, model-guided optimization can increase the likelihood of identifying viable formulations while reducing the number of experiments, development time, and quantity of valuable peptide and formulation materials required to reach a focused development region.

**Corbion and Intrepid Labs Capabilities**

Formulation transforms a drug into a viable medicine. A suboptimal formulation can contribute to clinical failure or result in an efficacious product whose patient acceptability, adherence, manufacturability, cost of goods, or supply chain requirements limit its real-world and commercial success. Yet early target product profiles often prioritize near-term performance measures while deferring objectives that influence patient use and commercialization. AI-driven design enables broader formulation spaces to be evaluated against a more complete target product profile from the outset, accelerating development while reducing the consumption of drug and formulation materials. The

| CORBION | INTREPID LABS |
|---|---|
| **Polymer and biomaterials capabilities** | **Formulation and automation capabilities** |
| PURASORB® bioresorbable polymers | AI/ML-enabled drug formulation design and multi-objective optimization |
| GMP-grade PLGA, PLA, and block copolymers | Automated formulation preparation and characterization |
| Materials designed for controlled-release drug delivery applications | Accelerated development with reduced API and material consumption |
| Custom polymer synthesis and scale-up support | Expertise across LAIs, liquid and solid oral, and parenteral formulations |

**AI-guided polymer and formulation design to accelerate the development of robust, scalable and patient-centered LAIs.**

resulting models learn composition-performance relationships and help define formulation boundaries, providing early insight into robustness and guiding further optimization through clinical development, scale-up, and manufacturing. Through the collaboration between Corbion and Intrepid Labs, an AI algorithm trained on data generated using commercially available and innovative Corbion materials is now available to the market to accelerate the design and optimization of polymer-based LAIs.

Supplementary information

# Accelerating the Development of PLGA In Situ Forming Depots Through AI-Driven Multi-Objective Optimization

Pauric Bannigan,† Siddarth Chandrasekaran,† Brigitte A. G. Lamers,‡ Inge Hermsen,‡ Gary Tom,† Riley J. Hickman,† Bahar Yeniad,‡ Morgan Fox,‡ Christine Allen†

*†Intrepid Labs Inc., 140 - 101 College St, Toronto, ON M5G 1L7. [E]:* christine@intrepidlabs.ai
*‡Corbion N.V., Arkelsedijk 46, Gorinchem, 4206 AC, [E]:* morgan.fox@corbion.com

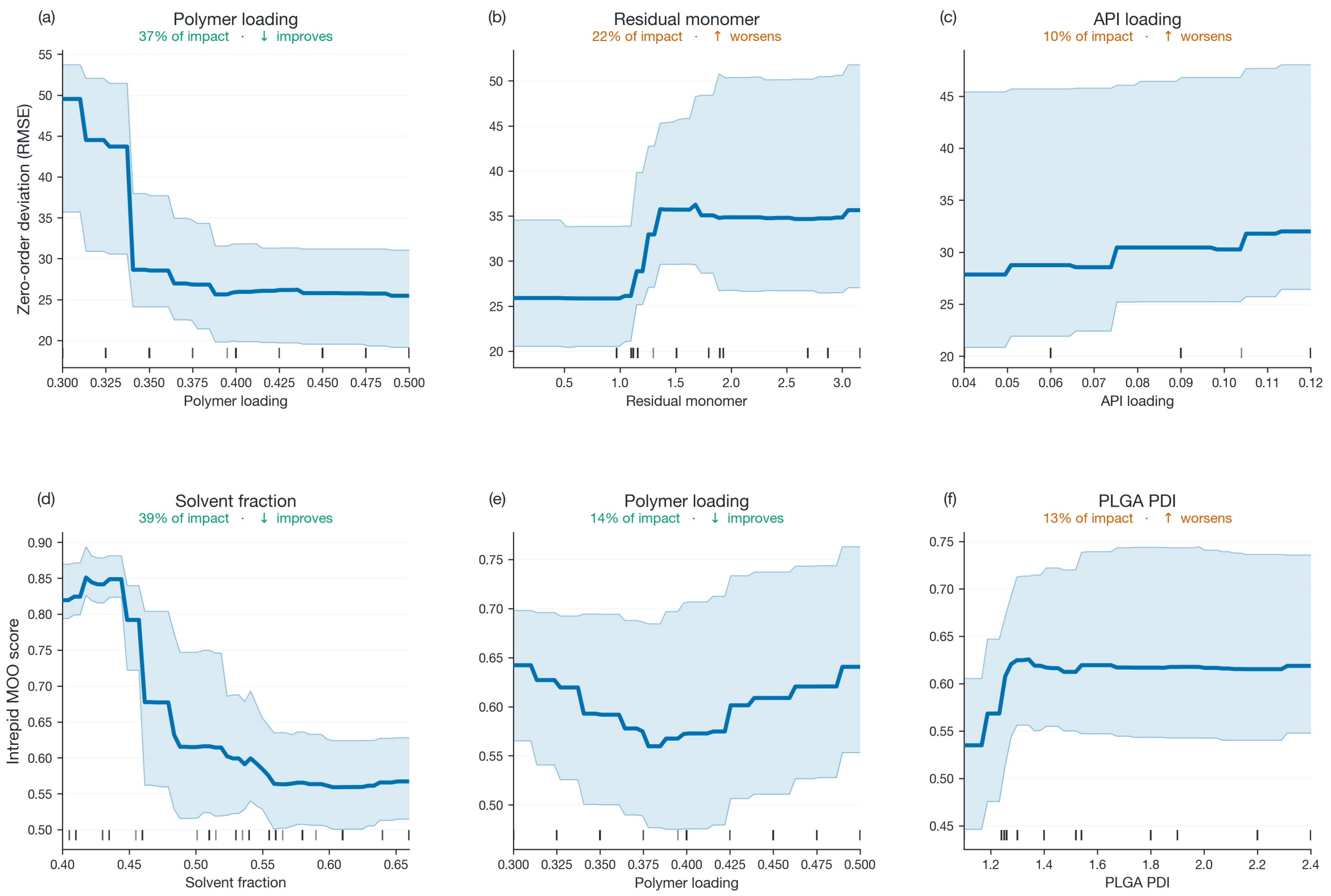

**Figure S1. Partial-dependence effect curves showing the modelled relationships between the three most influential formulation attributes and each of the four optimization objectives across the sampled design space.** Rows correspond to zero-order release deviation (a–c) and Intrepid multi-objective optimization score (d-f). For each panel, the indicated attribute was varied between its 5th and 95th percentile while all other attributes retained their measured values. Emulator predictions are summarised as the median response (solid line) and interquartile range (shaded band). Rug marks along the x-axis show the attribute values across experimental runs and indicate where the curves are supported by data. Lower values are preferred for all four objectives; descending curves therefore indicate improved performance, whereas ascending curves indicate worsening performance. The percentage above each panel represents the attribute's share of the total mean absolute SHAP contribution for that objective.

| *Strategy* | *F136* | *F139* | *F148* | *F162* | *At least one lead* |
|---|---|---|---|---|---|
| *ANDROMEDA 1* | 13 | 18 | 28 | 9 | 30 / 30 |
| *Random search* | 2 | 7 | 6 | 8 | 19 / 30 |
| *DoE* | 14 | 5 | 3 | 2 | 17 / 30 |

**Table S1. Frequency of lead formulation discovery across 30 independent simulated campaigns for each search strategy.** Values indicate the number of campaigns in which each lead formulation was sampled within a budget of 160 experiments. "At least one lead" indicates that one or more of F136, F139, F148, or F162 was sampled. F148 had the most favourable MOO score and was therefore the lead most directly aligned with the optimization objective. ANDROMEDA 1 sampled F148 in 28 of 30 campaigns, compared with 6 for random search and 3 for DoE, and sampled at least one lead in all 30 campaigns. ANDROMEDA 1 also sampled F139 more frequently than either comparator, while F136 was sampled with similar frequency by ANDROMEDA 1 and DoE.